\documentclass[runningheads]{llncs}
\usepackage[T1]{fontenc}
\usepackage{graphicx,verbatim}
\usepackage{amsmath}
\usepackage{amssymb}
\usepackage{float}
\usepackage{booktabs}
\usepackage{tikz}
\usepackage{graphicx}

\begin{document}
\title{On the Cone Effect and Modality Gap in Medical Vision–Language Embeddings}
\titlerunning{On the Cone Effect and Modality Gap in Medical VLMs}
\begin{comment}  %% Removed for anonymized MICCAI submission
\author{First Author\inst{1}\orcidID{0000-1111-2222-3333} \and
Second Author\inst{2,3}\orcidID{1111-2222-3333-4444} \and
Third Author\inst{3}\orcidID{2222--3333-4444-5555}}
%
\authorrunning{F. Author et al.}
\institute{Princeton University, Princeton NJ 08544, USA \and
Springer Heidelberg, Tiergartenstr. 17, 69121 Heidelberg, Germany
\email{lncs@springer.com}\\
\url{http://www.springer.com/gp/computer-science/lncs} \and
ABC Institute, Rupert-Karls-University Heidelberg, Heidelberg, Germany\\
\email{\{abc,lncs\}@uni-heidelberg.de}}

\end{comment}

\author{
David Restrepo\inst{1}\and
Miguel L Martins\inst{2}\and
Chenwei Wu\inst{3}\and
Luis Filipe Nakayama\inst{4}\and
Diego M. Lopez\inst{5}\and
Stergios Christodoulidis\inst{1}\and
Maria Vakalopoulou\inst{1}\and
Enzo Ferrante\inst{6}
}

\authorrunning{Restrepo et al.}

\institute{
CentraleSupélec, Université Paris-Saclay, France
%\email{\{david.restrepo, stergios.christodoulidis, maria.vakalopoulou\}@centralesupelec.fr}
\and
University of Porto, Portugal
%\email{miguelopesmartins@gmail.com}
\and
University of Michigan, USA
%\email{chenweiw@umich.edu}
\and
Federal University of São Paulo, Brazil
%\email{luisnaka@mit.edu}
\and
Universidad del Cauca, Colombia
%\email{dmlopez@unicauca.edu.co}
\and
Universidad de Buenos Aires, Argentina
%\email{eferrante@sinc.unl.edu.ar}
}
  
\maketitle              % typeset the header of the contribution
\begin{abstract}

Vision--Language Models (VLMs) exhibit a characteristic ``cone effect'' in which nonlinear encoders map embeddings into highly concentrated regions of the representation space, contributing to cross-modal separation known as the modality gap. While this phenomenon has been widely observed, its practical impact on supervised multimodal learning—particularly in medical domains—remains unclear. In this work, we introduce a lightweight post-hoc mechanism that keeps pretrained VLM encoders frozen while continuously controlling cross-modal separation through a single hyperparameter $\lambda$. This enables systematic analysis of how the modality gap affects downstream multimodal performance without expensive retraining. We evaluate generalist (CLIP, SigLIP) and medically specialized (BioMedCLIP, MedSigLIP) models across diverse medical and natural datasets in a supervised multimodal settings. Results consistently show that reducing excessive modality gap improves downstream performance, with medical datasets exhibiting stronger sensitivity to gap modulation; however, fully collapsing the gap is not always optimal, and intermediate, task-dependent separation yields the best results. These findings position the modality gap as a tunable property of multimodal representations rather than a quantity that should be universally minimized.

\end{abstract}
\section{Introduction}
Self-supervised learning via contrastive language-image pretraining (CLIP) has established itself as a cornerstone for developing frontier solutions in both natural and medical imaging domains \cite{radford2021learning,zhang2022contrastive}. By leveraging multi-modal self-supervision, this framework allows to capture complementary information from both modalities in a joint representation space. However, the contraction enforced by non-linearities in deep networks leads to the so called ``cone effect'' \cite{liang2022mind}, where representations are mapped to a very compact region in the representation space. Additionally, the contrastive learning loss optimization \cite{infonce} in CLIP models, introduces a repulsive component that clusters the representations into distant modality-specific regions of the representation manifold, a phenomenon called \emph{modality gap} \cite{liang2022mind}. 
It is still not clear if the existence of a domain gap is fundamentally harmful, although it is known that it can affect downstream performance \cite{liang2022mind}. Empirical and theoretical literature also suggest that CLIP alignment is also suboptimal in terms of intra and inter modality alignment \cite{liang2022mind,shi2023towards,zhang2023diagnosing,schrodi2024two,lee2025diffusion,mistretta2025cross}.

The simplest mechanism of action to correct potential misalignment is to manipulate the length of the domain-gap in the embedding space. However, from an information theoretic perspective, simply narrowing the gap as much as possible may lead to perfect inter-modality alignment, but comes at a potentially high information cost exclusive to each modality \cite{jiang2023understanding}. It is possible to minimize this gap in a more constrained fashion \cite{shi2023towards,hu2024reclip,eslami2024mitigate}, but current methods need to be integrated during the pre-training stage of the CLIP model itself. This is also the case for other approaches that address inter or intra-class alignment, relying on explicit terms in the pre-text objective \cite{jiang2023understanding,mistretta2025cross} or training that depends on model inversion methods \cite{huang2025mind,lee2025diffusion}.  

\begin{figure}[t]
    \centering
    \includegraphics[width=.9\textwidth]{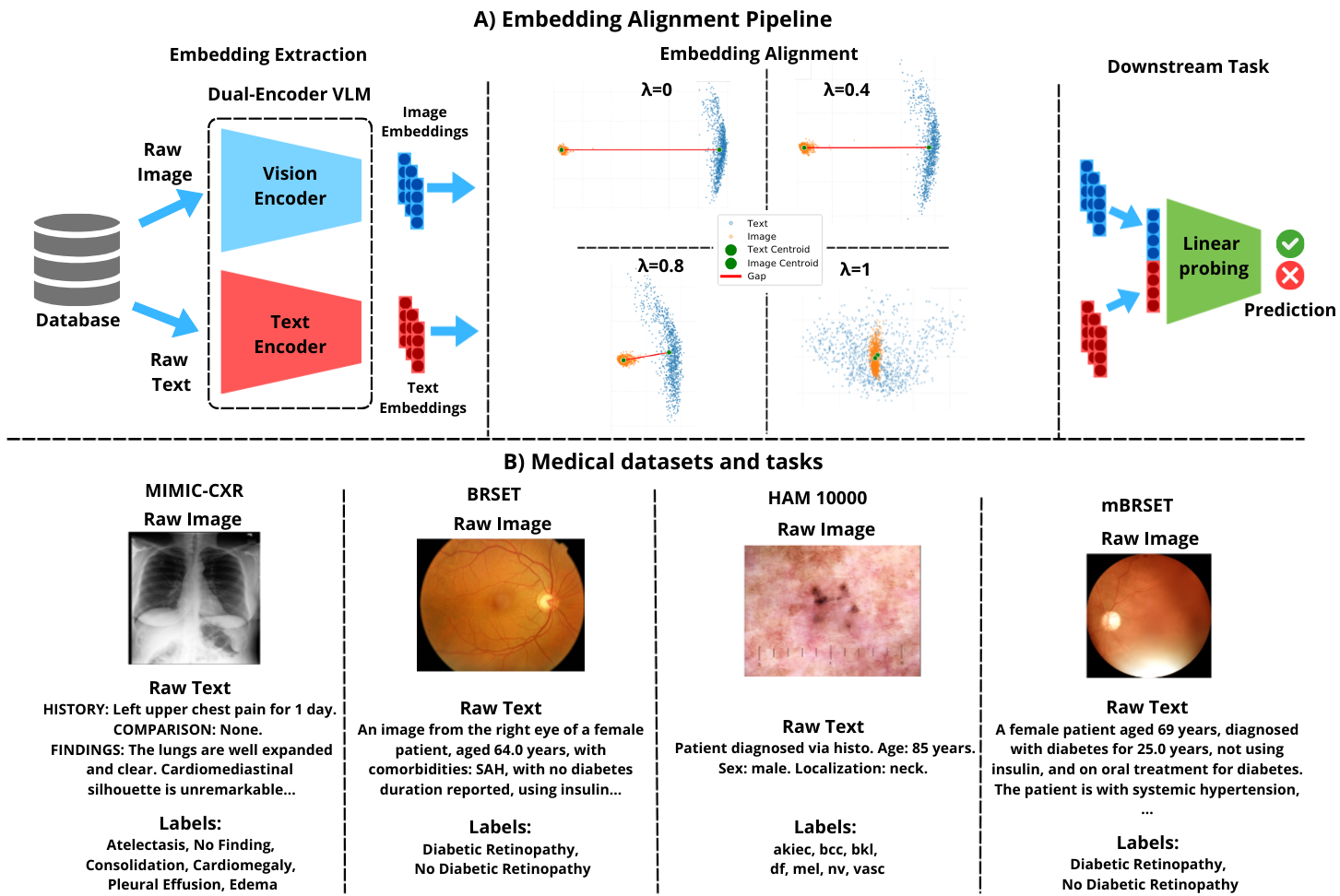}
    \caption{Overview of the proposed framework. (A) Embedding alignment pipeline: raw image and text features are extracted, shifted by a hyperparameter $\lambda$ to modulate the modality gap, and evaluated via linear probing. (B) Overview of medical datasets and corresponding multimodal tasks used in this study.}
    \label{fig:method_overview}
\end{figure}

In practice, CLIP-based models are require heavy computation and data for training \cite{radford2021learning}. Consequently, there is a critical need for lightweight mechanisms that can manipulate the modality gap post-hoc, adapting pre-trained representations to downstream requirements without expensive retraining. In contrast to prior work that primarily characterizes the modality gap in zero-shot settings \cite{liang2022mind}, we investigate its role in supervised multimodal learning where image and text signals can be complementary. We introduce a simple method controlled by a single hyperparameter $\lambda$, that continuously adjusts inter-modal separation without modifying the pretrained encoders as shown in figure \ref{fig:method_overview}-A. %Crucially, we perform a systematic analysis of downstream performance as a function of $\lambda$, revealing task- and domain-dependent optima that are particularly pronounced in medical data.

Our contributions are summarized as follows: \textit{(i)} we introduce a framework that keeps modality embedders frozen, and the domain-gap is selected as a function of the downstream task, \textit{(ii)} we perform an extensive analysis of the consequences of the cone effect and the domain gap on a collection of two medical and two natural VLMs pre-trained, across a variety of four medical and four natural datasets, \textit{(iii)} we study the impact of the proposed data-driven domain gap adaptation in supervised multi-modal downstream tasks, and its consequences in medical data. 

Our experiments indicate that the cone effect and modality gap are significantly more pronounced in medical datasets due to lower semantic and visual diversity compared to natural domains. We demonstrate that while narrowing this gap through our $\lambda$-alignment framework consistently improves downstream performance, the optimal separation is task-dependent, suggesting that medical VLMs require carefully tuned cross-modal overlap to preserve critical modality-specific information. Code and implementation details are available in the anonymous repository:
\url{https://anonymous.4open.science/r/Embedding-Alignment-Anonymized-7D2B/}.

% =========================
\section{Methods}
\label{sec:methods}

\subsection{Problem Statement: Cone Effect and Modality Gap in Vision--Language Models}
\label{sec:problem_theory}
\subsubsection{Modality Gap in Vision--Language Embedding Spaces.}
\label{sec:modality_gap}
Vision--Language Models (VLMs) such as CLIP \cite{radford2021learning} and SigLIP \cite{zhai2023sigmoid} learn joint embedding spaces by mapping images and texts into a shared representation. Let $f_\theta:\mathcal{I}\rightarrow\mathbb{R}^d$ and $g_\phi:\mathcal{T}\rightarrow\mathbb{R}^d$ denote the image and text encoders. We consider normalized embeddings
%\begin{equation}
$v=\frac{f_\theta(I)}{\|f_\theta(I)\|_2},
t=\frac{g_\phi(T)}{\|g_\phi(T)\|_2}.$
%\end{equation}
%By construction $v,t\in\mathbb{S}^{d-1}$, where the embeddings are first computed in $\mathbb{R}^d$ and then $\ell_2$-normalized, so they lie on the unit hypersphere $\mathbb{S}^{d-1}$. 
By construction $v,t$ live in the unit hypersphere $\mathbb{S}^{d-1}$ (i.e. $v,t\in\mathbb{S}^{d-1}$) since the embeddings are first computed in $\mathbb{R}^d$ and then $\ell_2$-normalized, which constrains them to have unit norm.
Despite contrastive training, embeddings from each modality are likely to be far apart in $S^{d-1}$; this problem is called the \emph{modality gap}~\cite{liang2022mind}. We quantify this gap through the centroid difference: %Since $u$ and $t$ likely map to dense regions on $S^{d-1}$ we use the expected value to compute the relative displacement: 

%\begin{equation}
%\mu_v=\mathbb{E}[v],
%\qquad
%\mu_t=\mathbb{E}[t],
%\qquad
%\Delta=\mu_v-\mu_t,
%\label{eq:gap_vector}
%\end{equation}

\begin{equation}
\mu_v = \frac{1}{N}\sum_{i=1}^{N} v_i,
\qquad
\mu_t = \frac{1}{N}\sum_{i=1}^{N} t_i,
\qquad
\Delta = \mu_v - \mu_t,
\label{eq:gap_vector}
\end{equation}

\noindent where $\|\Delta\|_2$ acts as proxy for the domain gap.

% ------------------------------------------------------------

\subsubsection{Cone Effect in Vision--Language Embedding Spaces.}
\label{sec:cone_effect}

The cone effect emerges from the compositional nonlinearities inherent in deep models, where embeddings concentrate within narrow angular regions of the hypersphere~\cite{liang2022mind}. This holds for all non-linear activations. Liang et al.~\cite{liang2022mind} show that for the specific case of ReLU activations, $\phi(x)=\max(x,0)$, which restricts the output space. %, monotonically increasing cosine similarity with the depth of the network,
%\begin{equation}
%\cos(\phi(Wx_1),\phi(Wx_2)) \geq \cos(x_1,x_2),
%\label{eq:cone_similarity}
%\end{equation}.
Additionally, for a mini-batch of size $B$, the InfoNCE \cite{infonce} loss within the context of CLIP can be understood as the composition of the attractive/repulsive forces:

\begin{equation}
\mathbb{E}\left[-\log\frac{\exp(v_i^\top t_i/\tau)}
{\sum_{j=1}^B\exp(v_i^\top t_j/\tau)}\right]
=
\underbrace{-\frac{1}{\tau}v_i^\top t_i}_{\text{attraction}}
+
\underbrace{\log\sum_{j=1}^B\exp(v_i^\top t_j/\tau)}_{\text{repulsion}},
\label{eq:clip_attract_repel}
\end{equation}

\noindent where the first term pulls positive pairs closer, while the second term is a \emph{repulsive} term with respect to the negative pairs in the mini-batch.
\subsubsection{Cone effect in Medical Domains.}
\label{sec:medical_specifics}
Since data in medical domain $U_{\text{med}}$ are more homogeneous than in natural images, it is reasonable to suppose that the cone effect will manifest in a cap of smaller radius in the representation space when compared with natural domain data, $U_{\text{nat}}$.
Let $h_\Theta(U)$ denote the output of a neural network parameterized by $\Theta$. Then the total output variance satisfies the law of total variance:
\begin{equation}
\mathrm{Var}\!\left[h_\Theta(U)\right]
=
\underbrace{\mathbb{E}_\Theta\!\left[\mathrm{Var}\!\left(h_\Theta(U)\mid\Theta\right)\right]}_{\text{data-induced variance}}
+
\underbrace{\mathrm{Var}_\Theta\!\left(\mathbb{E}\!\left[h_\Theta(U)\mid\Theta\right]\right)}_{\text{weight-induced variance}}.
\label{eq:total_variance_liang}
\end{equation}

In medical datasets, one expects reduced semantic and visual diversity, resulting in lower variance in the data-dependent term,
\begin{equation}
\mathbb{E}_\Theta\!\left[\mathrm{Var}\!\left(h_\Theta(U_{\text{med}})\mid\Theta\right)\right]
\leq
\mathbb{E}_\Theta\!\left[\mathrm{Var}\!\left(h_\Theta(U_{\text{nat}})\mid\Theta\right)\right],
\label{eq:data_variance_medical}
\end{equation}
making medical representations even more susceptible to the cone effect. 
However, backbones pre-trained on medical data introduce domain-specific projections through specialized pretraining. Such pre-training affects the weight-induced term in Eq.~\ref{eq:total_variance_liang}, which we conjecture results in a lower-dimension manifold that depends on the main directions of variation of the medical data. 
\subsubsection{Cross-domain Embedding Alignment Intervention.}
\label{sec:alignment}
To mitigate the modality gap without modifying encoder weights, we apply an inference-time embedding alignment intervention. For $\lambda\in[0,1]$, embeddings are shifted along the gap vector and renormalized:
\begin{equation}
t'=\frac{t+\frac{\lambda}{2}\Delta}{\|t+\frac{\lambda}{2}\Delta \|_2},
\qquad
v'=\frac{v-\frac{\lambda}{2}\Delta}{\|v-\frac{\lambda}{2}\Delta \|_2}.
\label{eq:alignment}
\end{equation}
By varying $\lambda$, we control the size of the domain gap.
% ============================================================
\subsection{Experimental Setup}
\label{sec:experimental_setup}

\subsubsection{Vision--Language Backbones}
\label{sec:backbones}
We evaluate both general and medical-domain VLMs. Specifically, we consider CLIP (ViT-B/32) \cite{radford2021learning} and SigLIP (ViT-B/16 \cite{zhai2023sigmoid} as natural-domain baselines, and BioMedCLIP \cite{zhang2023biomedclip} and MedSigLIP \cite{sellergren2025medgemma} as medical-domain variants trained on biomedical corpora and clinically curated image--text pairs. For each backbone, we extract frozen image and text embeddings independently using Eq.~(1), enabling direct geometric comparison across modalities and domains without confounding effects from downstream fine-tuning.
% ------------------------------------------------------------
\subsubsection{Datasets and Domain Shift Regimes}
\label{sec:datasets}

We conduct experiments on four medical benchmarks—MIMIC-CXR~\cite{johnson2019mimic}, comprising chest X-rays paired with radiology reports for multilabel classification using CheXpert labels~\cite{irvin2019chexpert}: Atelectasis, Cardiomegaly, Consolidation, Edema, Pleural Effusion; HAM10000~\cite{tschandl2018ham10000}, containing dermatoscopic skin lesion images and metadata represented as text for 7-class, multi-class classification; BRSET~\cite{nakayama2024brset} and mBRSET~\cite{nakayama2024mbrset}, which provide retinal fundus images with clinical and demographic metadata for binary diabetic retinopathy prediction. Additionally, we include four natural-domain datasets as reference, such as Recipes5k~\cite{bolanos2017food} (food images with ingredient lists for multi-class classification), DAQUAR~\cite{malinowski2014multi} (indoor scene VQA multi-label), COCO-QA~\cite{ren2015exploring} (MS-COCO VQA multi-class), and Fakeddit~\cite{nakamura2020r} (multimodal fake news binary classification).

Each dataset consists of paired image--text samples with labels $y_i$ corresponding to either multi-class or multi-label prediction tasks as shown in Fig. \ref{fig:method_overview}-B. %Datasets are partitioned into training, validation, and test splits.

\subsubsection{Embedding Alignment Protocol}
\label{sec:lambda_protocol}

For each dataset and backbone, the modality gap vector $\Delta$ is estimated from the pre-trained embeddings as
\begin{equation}
\Delta = \mu_v - \mu_t,
\qquad
\mu_v=\mathbb{E}_{(I,T)\sim \mathcal{D}^{\text{train}}}[v],
\quad
\mu_t=\mathbb{E}_{(I,T)\sim \mathcal{D}^{\text{train}}}[t].
\end{equation}

We then compute the aligned embeddings $\{(v'_i,t'_i)\}$ for $\lambda\in\{0.0,0.1,\dots,1.0\}$, adjusting the domain gap, $\|\Delta\|_2$.
% ------------------------------------------------------------
\subsubsection{Linear Probe Downstream Evaluation}
\label{sec:linear_probe_eval}
To isolate embedding quality from nonlinear classifier capacity, we evaluate aligned representations using an Early Fusion Linear Probe. For each aligned pair $(v'_i,t'_i)$, we form the concatenated feature vector
and input into a model with no hidden layers or nonlinear activations.
%Probe parameters $(W,b)$ are optimized on $\mathcal{D}_k^{\text{train}}$ by minimizing %the empirical risk
%\begin{equation}
%\min_{W,b}\;
%\frac{1}{|\mathcal{D}_k^{\text{train}}|}
%\sum_{(x_i,y_i)\in\mathcal{D}_k^{\text{train}}}
%\ell(\hat{y}_i,y_i),
%\label{eq:erm_probe}
%\end{equation}
The models were trained using class-weighted binary cross-entropy for multi-label datasets and categorical cross-entropy for multi-class datasets. 
Optimization was performed using stochastic gradient descent with momentum $0.9$ and learning rate $3\times 10^{-4}$, with early stopping based on validation loss.
% ------------------------------------------------------------

\subsubsection{Evaluation Metrics and Statistical Protocol}
\label{sec:metrics_protocol}

Model performance is assessed on held-out test splits $\mathcal{D}^{\text{test}}$ using the Area Under the Receiver Operating Characteristic Curve (AUC). For each backbone, dataset, and alignment strength $\lambda$, experiments are repeated across five independent random seeds. Final results are reported as mean $\pm$ standard deviation.
%\begin{equation}
%\mathrm{AUC}_{k,\lambda} =
%\frac{1}{5}\sum_{s=1}^5 \mathrm{AUC}_{k,\lambda}^{(s)}
%\;\;\pm\;\;
%\mathrm{Std}\big(\mathrm{AUC}_{k,\lambda}^{(s)}\big),
%\end{equation}
% ------------------------------------------------------------
\subsubsection{Modality Gap Visualization}
\label{sec:geometry_vis}

To complement the evaluation, we visualize modality cones and centroid displacement using Principal Component Analysis (PCA). Normalized embeddings are projected into low-dimensional subspaces, allowing qualitative inspection of (i) cross-modal separation induced by $\Delta$, and (ii) intra-modality angular concentration associated with the cone effect.

To quantify the embedding cluster concentration (cone effect), we use the Mean Resultant Length ($R$),
\begin{equation}
R \;=\; \left\|\frac{1}{N}\sum_{i=1}^{N} v_i \right\|_2.
\end{equation}

Intuitively, $R$ measures how aligned the embeddings are in direction: if vectors are uniformly spread on the unit hypersphere, their mean cancels out and $R\approx 0$ (isotropic distribution), whereas if they concentrate within a narrow cone, their directions reinforce and $R\rightarrow 1$. Thus, larger $R$ indicates stronger anisotropy and tighter modality cones.

\section{Results and Discussion}
\label{sec:results_discussion}

\begin{figure}[h]
    \centering
    \includegraphics[width=.9\linewidth]{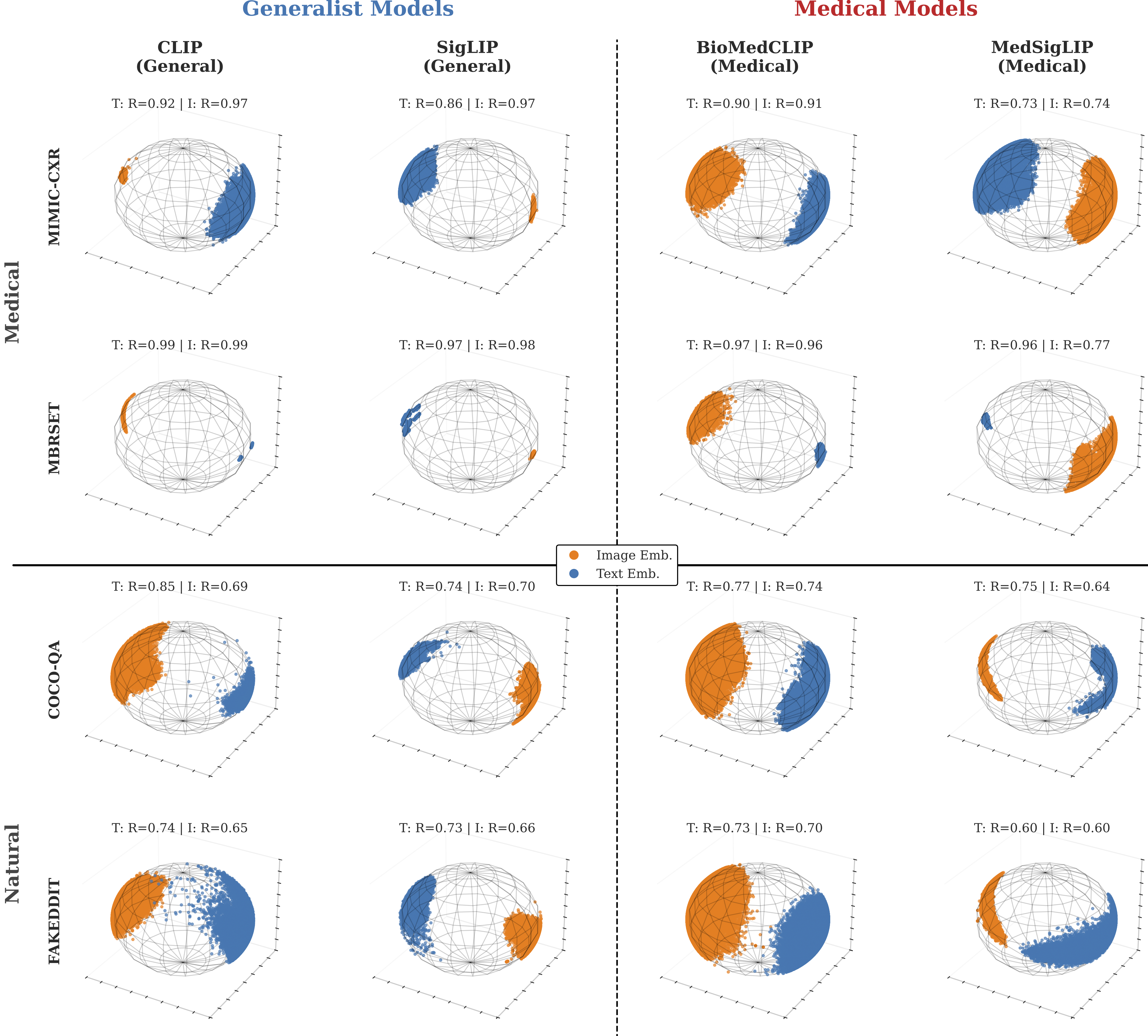}
    \caption{Low-dimensional visualization of image and text embeddings across datasets and backbones. Projections are shown using the first two principal components in three-dimensional ambient space.}
    \label{fig:geometry}
\end{figure}

\begin{table}[h]
\centering
\caption{Mean Resultant Length ($R$), quantifying the degree of embedding concentration for Text and Image. Higher values indicate narrower cones (stronger concentration). $R \in [0, 1]$.}
\resizebox
{\textwidth}{!}{%
\begin{tabular}{lcccc c cccc}
\toprule
 & \multicolumn{4}{c}{\textbf{Text Embeddings ($R$)}} & & \multicolumn{4}{c}{\textbf{Image Embeddings ($R$)}} \\
\cmidrule{2-5} \cmidrule{7-10}
\textbf{Dataset} & \textbf{CLIP} & \textbf{SigLIP} & \textbf{BioMed} & \textbf{MedSig} & & \textbf{CLIP} & \textbf{SigLIP} & \textbf{BioMed} & \textbf{MedSig} \\
\midrule
\multicolumn{10}{l}{\textit{Medical Datasets}} \\
\midrule
MIMIC-CXR & 0.915 & 0.862 & 0.899 & 0.726 & & 0.970 & 0.974 & 0.910 & 0.735 \\
HAM10000 & 0.919 & 0.809 & 0.889 & 0.970 & & 0.920 & 0.933 & 0.941 & 0.858 \\
BRSET & 0.985 & 0.956 & 0.990 & 0.980 & & 0.986 & 0.980 & 0.964 & 0.731 \\
mBRSET & 0.990 & 0.973 & 0.975 & 0.957 & & 0.986 & 0.982 & 0.958 & 0.770 \\
\midrule
\multicolumn{10}{l}{\textit{Natural Datasets}} \\
\midrule
COCO-QA & 0.852 & 0.743 & 0.769 & 0.751 & & 0.691 & 0.702 & 0.739 & 0.638 \\
Fakeddit & 0.745 & 0.730 & 0.734 & 0.604 & & 0.654 & 0.657 & 0.702 & 0.597 \\
Recipes5k & 0.775 & 0.760 & 0.869 & 0.632 & & 0.810 & 0.791 & 0.807 & 0.738 \\
DAQUAR & 0.903 & 0.793 & 0.816 & 0.881 & & 0.862 & 0.842 & 0.855 & 0.790 \\
\bottomrule
\end{tabular}}
\label{tab:cone_strength}
\end{table}

Figure~\ref{fig:geometry} visualizes the joint embedding geometry for a representative subset of datasets and backbones. We observe that across all models, both image and text representations concentrate into narrow angular regions of the hypersphere. The full quantitative analysis of this phenomenon is also available in Table~\ref{tab:cone_strength}, reporting the Mean Resultant Length ($R$) for all studied datasets. The results confirm that in medical datasets, the reduced semantic diversity yields tighter representations and consistently higher $R$ values compared to natural datasets.

We observe a persistent centroid displacement between modalities, with the modality gap largely aligned with the principal direction. Medical-domain backbones (BioMedCLIP, MedSigLIP) exhibit smaller cross-modal separation than generalist models for all the datasets, suggesting that medical pretraining can alleviate the cone effect for both medical, and non-medical datasets.

\begin{figure}[H]
    \centering
    \includegraphics[width=.9\linewidth]{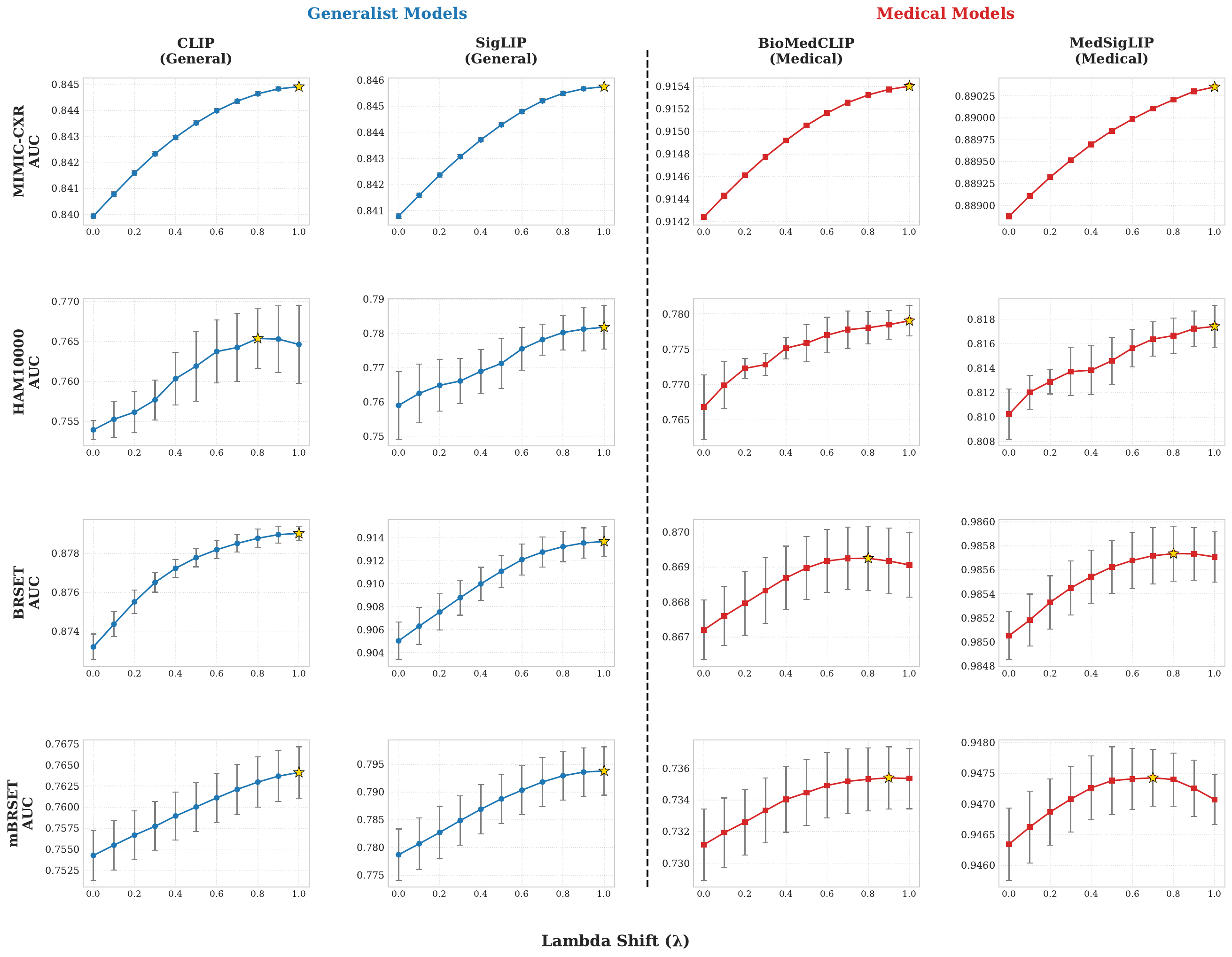}
    \caption{Downstream AUC as a function of alignment strength $\lambda$ across datasets and backbones. Moderate reduction of the modality gap consistently improves performance, while excessive closure can lead to saturation or degradation, particularly for medical-domain models. Error bars denote standard deviation across seeds.}
    \label{fig:lambda}
\end{figure}

Figure~\ref{fig:lambda} evaluates the downstream impact of controlled gap reduction via the alignment strength $\lambda$. Moderate closure of the modality gap consistently improves linear-probe AUC across datasets, indicating that excessive cross-modal displacement is detrimental. However, performance in some cases slightly degrades as $\lambda\!\rightarrow\!1$, particularly for medical-domain models on BRSET and mBRSET. This highlights that full collapse of the gap can remove modality-specific information, supporting the view that the optimal geometry corresponds to an intermediate, task-dependent separation rather than perfect overlap \cite{jiang2023understanding}. Generalist models benefit more strongly from post-hoc alignment on medical data, consistent with larger initial inter-modality misalignment under domain shift, while medical-tuned models show smaller but stable gains.

\section{Conclusion}
%We effectively demonstrated that the side effects of the cone-effect are pervasive in medical modalities, specially when using general purpose VLMs. Interestingly enough, we observed that VLMs seem to model allow more variance in their representation space for medical domain, while maintaining the same overall global structure for natural domains.

Our experiments in a broad set of medical and computer vision tasks effectively shown that the gap between modalities in representation space is far from optimal when optimizing a CLIP-like objective. 

We introduce a simple hyper-parameter, $\lambda$, that effectively allows narrowing the gap between the inter-domain embeddings during the downstream tasks using the original VLMs in inference mode. %This means that we do not require reformulating the overall pre-training framework, and thus the proposed method is significantly cheaper from an operational and resource perspective. % since it does not require pre-training of the backbones. 

Remarkably, when comparing to the baseline VLM gap ($\lambda=0$), our experiments suggest that closing the gap (i.e., increasing $\lambda$) is always desirable. For certain pairs of dataset and models, we observe slight dips in performance as $\lambda \rightarrow 1$. We contend this in an effect of the loss of modality-specific information in favor of cross-modality alignment. Our plan is to study the limiting behaviour $\lambda \rightarrow 1$ in future work by relating the domain-gap, the mutual information of the inter-domain representations, and the target variable of the downstream task. 
\begin{credits}
\subsubsection{\ackname} This project was supported by the European Union’s Horizon Europe research and innovation programme under the Marie Skłodowska-Curie COFUND grant agreement No 101127936 (DeMythif.AI). This project was supported by France 2030 funding, managed by the National Research Agency (ANR), as part of IA CLUSTER program, reference ANR-23-IACL-0003 - DATAIA CLUSTER. This work was granted access to the HPC resources of the Jean Zay supercomputer operated by IDRIS (CNRS) and to the Ruche Mesocentre of Université Paris-Saclay.
\end{credits}

\end{document}